\pdfoutput=1

\documentclass[11pt]{article}

\usepackage[]{acl}

\usepackage{times}
\usepackage{latexsym}

\usepackage[T1]{fontenc}

\usepackage[utf8]{inputenc}

\usepackage{microtype}

\usepackage{enumitem}
\usepackage{bookmark}
\usepackage{bm}
\usepackage{amsfonts}
\usepackage{subfigure}
\usepackage{graphicx}
\usepackage{booktabs}
\usepackage{makecell}
\usepackage{xcolor}
\usepackage{amsmath}
\usepackage{arydshln}
\usepackage{multirow}
\usepackage{stmaryrd}
\usepackage{color}
\usepackage{xspace}
\usepackage{makecell}
\usepackage{multicol}

\newcommand{\secref}[1]{\S\ref{#1}}

\newcommand{\ourmodel}{IBR\xspace}

\title{Interpretable Proof Generation via Iterative Backward Reasoning}

\author{ Hanhao Qu\textsuperscript{1} \quad Yu Cao\textsuperscript{3}\quad Jun Gao\textsuperscript{1}\quad Liang Ding\textsuperscript{3,4}\quad Ruifeng Xu\textsuperscript{1,2}\thanks{\;\;Corresponding author}\\
 \textsuperscript{1}Harbin Institute of Technology, Shenzhen\quad \textsuperscript{2}Peng Cheng Laboratory\quad  \\
\normalsize \texttt{\{hhqu0917,jgao95\}@stu.hit.edu.cn}\quad \texttt{xuruifeng@hit.edu.cn}\\
\textsuperscript{3}The University of Sydney, Australia \quad \textsuperscript{4}JD Explore Academy\\
\normalsize \texttt{\{ycao8647,ldin3097\}@uni.sydney.edu.au}\\
}

\begin{document}
\maketitle
\begin{abstract}
We present \ourmodel, an Iterative Backward Reasoning model to solve the proof generation tasks on rule-based Question Answering (QA), where models are required to reason over a series of textual rules and facts to find out the related proof path and derive the final answer.
We handle the limitations of existed works in two folds: 1) enhance the interpretability of reasoning procedures with detailed tracking, by predicting nodes and edges in the proof path iteratively backward from the question; 2) promote the efficiency and accuracy via reasoning on the elaborate representations of nodes and history paths, without any intermediate texts that may introduce external noise during proof generation.
There are three main modules in \ourmodel, \emph{QA and proof strategy prediction} to obtain the answer and offer guidance for the following procedure; \emph{parent node prediction} to determine a node in the existing proof that a new child node will link to; \emph{child node prediction} to find out which new node will be added to the proof.
Experiments on both synthetic and paraphrased datasets demonstrate that \ourmodel has better in-domain performance as well as cross-domain transferability than several strong baselines. Our code and models are available at \url{https://github.com/find-knowledge/IBR}.
\end{abstract}

\section{Introduction}
Endowing machines with reasoning capabilities is a longstanding problem~\citep{Newell} in the field of AI. Though existing tasks such as
multi-hop QA~\cite{hotpotqa,qangaroo} or logical-reasoning QA~\cite{reclor,drop} impose a higher requirement on the reasoning capabilities, they usually just request for an answer without the reasoning procedure that would make it interpretable.
Recently, \newcite{clark_synthetic_dataset} proposed new datasets and tasks for interpretable reasoning. Given a question, coupling with a set of facts (plain statements) and rules (implication relationships) that are expressed in natural language, there are two tasks: 1) predicting the binary answer; 2) generating the proof path behind this answer. 
Large-scale pretrained models have shown strong performance on the first subtask in the early work~\citep{roberta}, but there still remain challenges for the second one. These proof paths are usually more complicated than those involved in multi-hop QA tasks, as there are more nodes and branches rather than a single-directed chain.
\begin{figure}[t]
\setlength{\belowcaptionskip}{-0.2cm}
    \includegraphics[width=\linewidth]{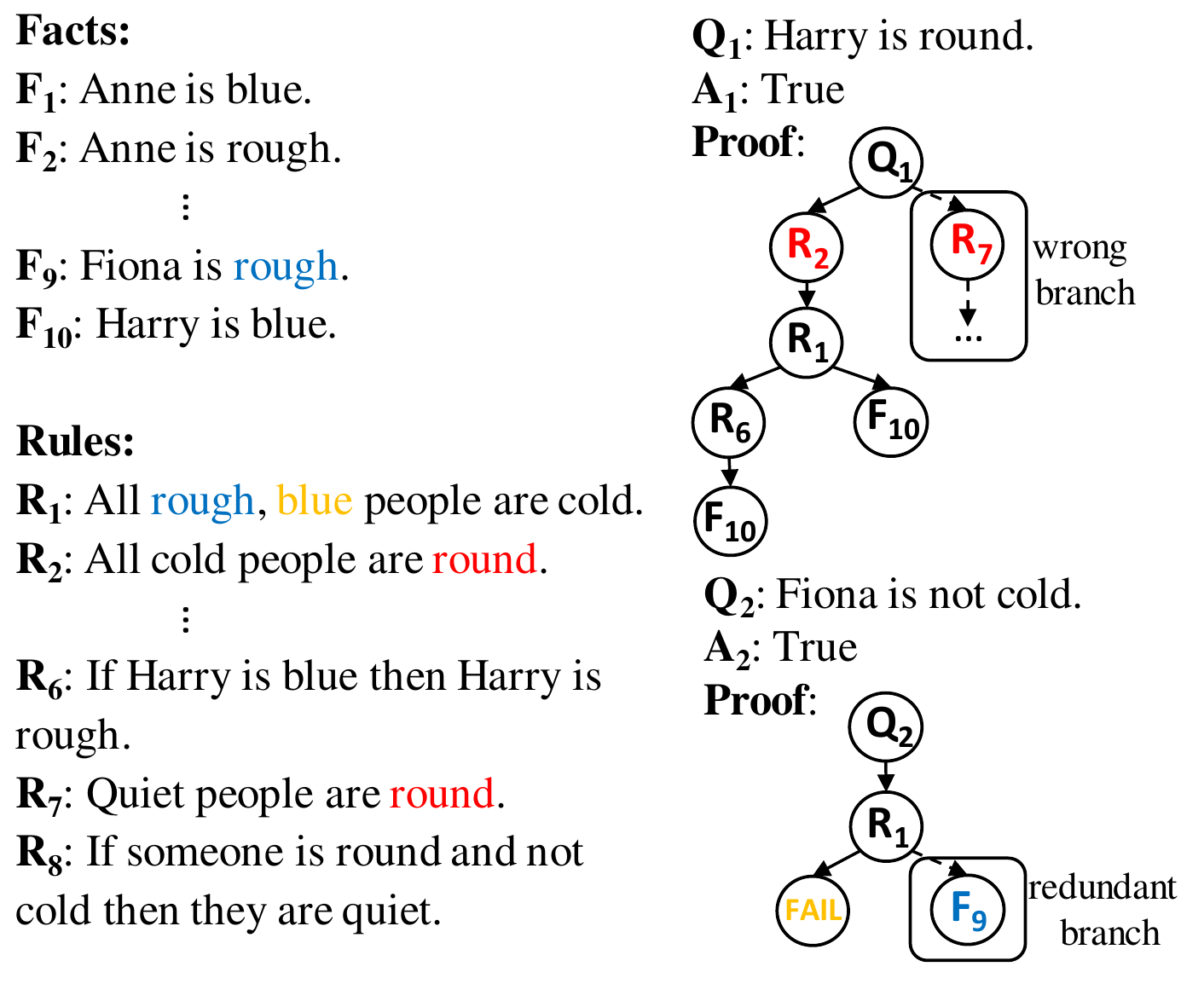}
    \caption{Illustration of generating proof iteratively. Regarding the proof path as a graph, and using the question as the initial node, other nodes and edges will be added step by step. (The gold proof is the obtained path in a reverse order exclude the question). The main challenges are wrong (cannot derive the answer) or redundant (can derive the answer, but the path is longer than the optimal one) branches may be involved. 
    }
    \label{fig:main_example}
\end{figure}

Several approaches have been proposed to simultaneously address the two subtasks.
PROVER~\citep{prover} and PROBR~\citep{probr} try to construct the reasoning path at once, where two classifiers are used to determine whether each node or edge is involved in the proof path respectively based on corresponding encoded representations. But they lack interpretability on tracking the detailed reason for selecting each step. To make proof generation more interpretable, Proofwriter~\citep{proofwriter} and EVR~\citep{evr} decompose complex reasoning over the question into multiple simple procedures, resulting in iterative and interpretable processes with the help of intermediate texts. Nevertheless, both of them suffer from efficiency and external errors issues. The reason is that they both require a large searching space, as they perform on the whole inferable texts and ignore the structure information from the history path that has been obtained. Moreover, the generation of intermediate text is costly and may introduce extra noise propagation.

Inspired by the top-down AMR parsing~\cite{cai_amr}, where a sentence is divided into sub-meanings iteratively, we present \textbf{I}terative \textbf{B}ackward \textbf{R}easoning (IBR) for better proof generation. It generates a proof path iteratively starting from the core component for QA, i.e. the question, making the process interpretable with trackable intermediate states.
Regarding a higher efficiency and accuracy, and two challenges mentioned in Figure~\ref{fig:main_example}, the proof generation module of \ourmodel simplifies the intermediate process of reasoning as well as avoids the unnecessary search for a possible unsuitable branch.
To add a new node and edge to the path, there are two steps in IBR for each iteration: 1) finding out the next parent node, i.e. one existing rule or fact in the parsed history path that a new node will become its child; 2) determine which rule or fact that will be the new child node and added to the path. Equipped with question-aware representations from a pre-trained encoder, along with structure-aware node and path features, our model can choose the optimal endpoint. It accomplishes reasoning with the highest possibility to obtain a correct subsequent proof path based on relevant features, getting rid of intermediate texts while avoiding redundancy on all possible texts than previous iterative works.

In addition, to make \ourmodel applicable for samples with incomplete proof paths, which are abandoned in the former backward iterative model EVR~\cite{evr}, we employ a proof strategy predictor to output a proof type. This prediction is then integrated into the later proof generation actions, making the process more controllable under different conditions. 

We validate our approach on several datasets that are widely used in previous studies (i.e. DU0-DU5, Birds-Electricity, and ParaRules) spanning different settings (i.e. fully-supervised, fewer training data, and out-of-domain). Experimental results show that, compared to existing strong baselines including both non-iterative and iterative ones, \ourmodel can achieve the best overall performance of proof generation and comparable answer prediction accuracy, along with noticeable generalization capability. 
Extensive analyses show that 1) the improvements come from our elaborately designed iterative and simplified proof generation modules, and 2) both the reasoning ability and latency could be significantly improved compared to former iterative models, making a better trade-off considering its reasonable interpretability.



\section{Related Work}

\paragraph{Question answering and reasoning.} Endowing machines to do reasoning over explicit knowledge is a primitive task~\citep{Newell}. Early works tried to solve it by converting texts into logic forms~\cite{Newell,Musen1988OfBA}. But such kinds of approaches can be affected by the error propagation caused by semantic parsing~\cite{zettlemoyer2005learning,Berant2013SemanticPO,Berant2014SemanticPV}.

Lately, question answering (QA) is employed as an important task for machine reasoning. Numerous datasets were proposed, including synthesized data~\cite{weston_babi}, comprehension on natural texts~\cite{rajpurkar_squad,joshi_triviaqa,fisch_mrqa} or more complex relationship reasoning~\cite{tafjord_quartz,lin_reasoning}. There are also multi-hop QA tasks like HotpotQA~\cite{hotpotqa} or QAngaroo~\cite{qangaroo}, and logical QA datasets such as ReClor~\cite{reclor} and LogiQA~\cite{liu_logiqa}, in which textual rules need to be inferred implicitly from a long supporting context. Plenty of studies try to solve these problems via neural networks and achieve remarkable performance~\cite{joshi_spanbert,yu_qanet,shao_graph}. Nevertheless, nearly all of them only focus on the prediction of final answers and neglect the acquisition of interpretable proofs. Although some datasets provide proof paths for better interpretability, these paths are only short chains with very few entities and cannot teach models to generate complex proofs.

\paragraph{Proof generation.}
NLProlog~\cite{nlprolog} first employs logic programming to search for a proof and then predicts the answer in multi-hop QA.
Recently, \citet{clark_synthetic_dataset} propose new rule-based QA datasets for this line of research that include more complex proof paths, and present RuleTaker to answer questions. \citet{prover} argue that producing answer proofs makes models more reliable and propose PROVER, a transformer-based model that enumerates all possible nodes and edges of a proof path and predicts whether each one exists at once based on their embeddings. PROBR~\cite{probr} further improves this framework using the probabilistic graph to model more variables.  
There has been also an increasing interest in solving proof generation iteratively. EVR~\cite{evr} splits the question into sub-questions, using generated intermediate texts to guide proof generation step by step. ProofWriter~\cite{proofwriter} shares a similar idea but uses intermediate textual conclusions instead and a more powerful T5-11B model~\cite{t5} for generation, which makes it hard to reproduce. 
\ourmodel is also an iterative model, being more interpretable than at-once models. Despite getting rid of intermediate texts and directly using various representations to finish each step, it improves efficiency and effectiveness.

\section{Methodology}

\subsection{Task Definition}
\label{sec:task_definition}

\begin{figure}[t]
\setlength{\belowcaptionskip}{-0.2cm}
    \centering
    \includegraphics[width=1\linewidth]{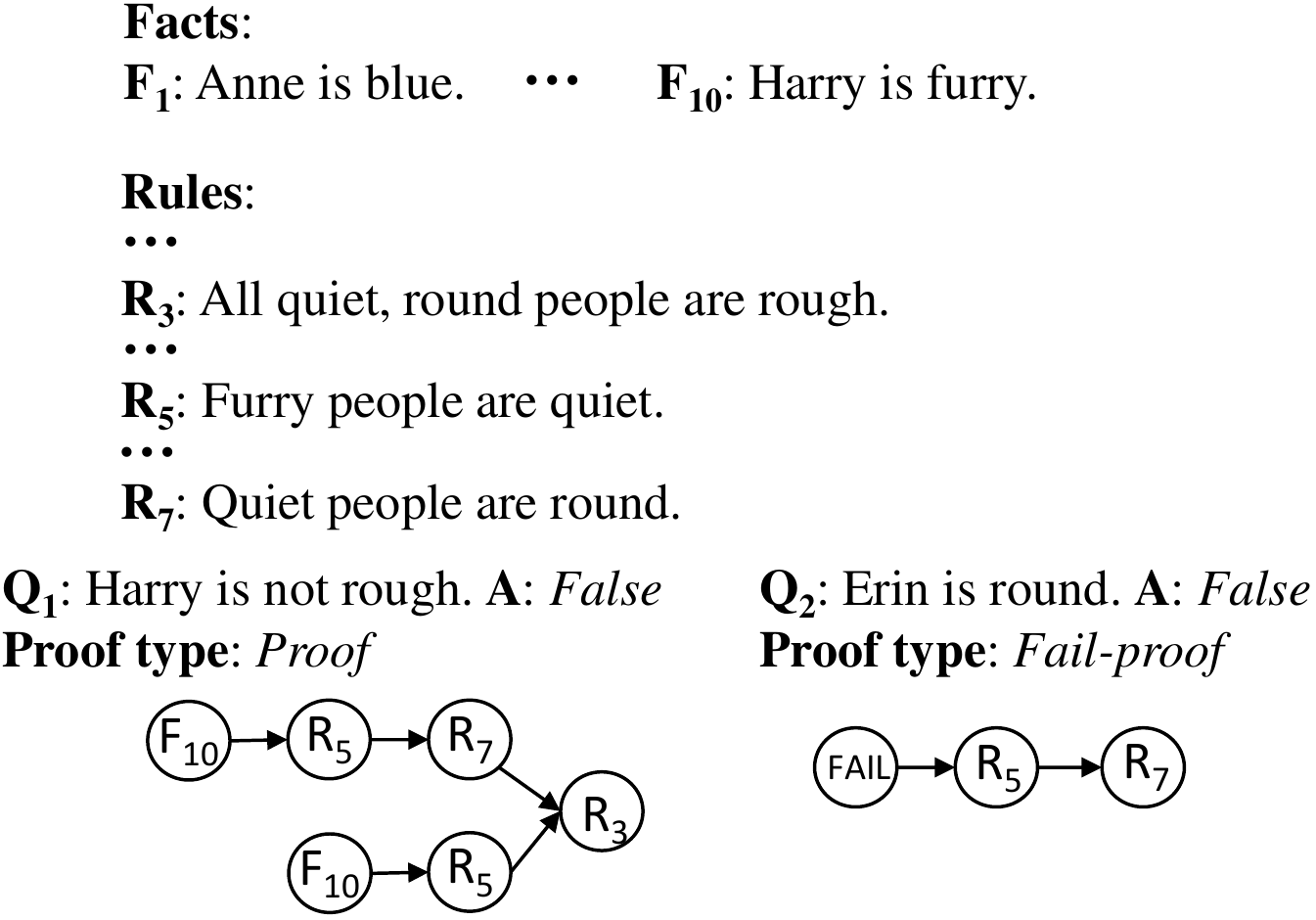}
    \caption{Examples of \textit{Proof} and \textit{Fail-proof} strategies.}
    \label{fig:proof_example}
\end{figure}

We first formulate the proof generation task as follows.
Given a tuple $(C,Q,A,P)$, where $C=\{RF_i\}$ is the contexts containing several textual rules and facts $RF$, $Q$ is the question, $A \in \{$\textit{True}, \textit{False}$\}$ is the answer, and $P$ indicates the proof path for the detailed reasoning procedure to derive $A$, our goal is twofold: 1) predicting the answer $A$, and 2) generating the proof path $P$.
Taking \textbf{DU0-DU5}~\cite{clark_synthetic_dataset} dataset as example, $P$ is a single-directed acyclic graph having the shortest path to derive $A$. $P$ can start from one or multiple nodes but must end in one node that directly entails or contradicts $Q$. A node in $P$ can be a fact, a rule, or a special NAF (Negation As Failure) node\footnote{A start node when the negation condition in the next node has no corresponding fact nor rule node, and the negation will be considered as true. E.g., there is no item in $C$ related to ``\emph{Anne is big}'', its negation ``\emph{Anne is not big}'' will be considered as true.}. Edges between nodes indicate that the start nodes can be used to prove the end nodes during reasoning. Proofs in the dataset can be roughly classified into two types according to their \textbf{strategies} $S$ to prove the question: (1)\emph{Proof}: the question can be directly proven to be True or False using the given $C$ and NAF; (2) \emph{Fail-Proof}: the question cannot be explicitly deduced barely using $C$ and NAF as some key information is missed, hence a positive statement is judged as \textbf{False} while a negative statement as \textbf{True} in such cases (Figure~\ref{fig:proof_example}).



\begin{figure*}[t]
\setlength{\belowcaptionskip}{-0.2cm}
    \centering
    \includegraphics[width=0.98\linewidth]{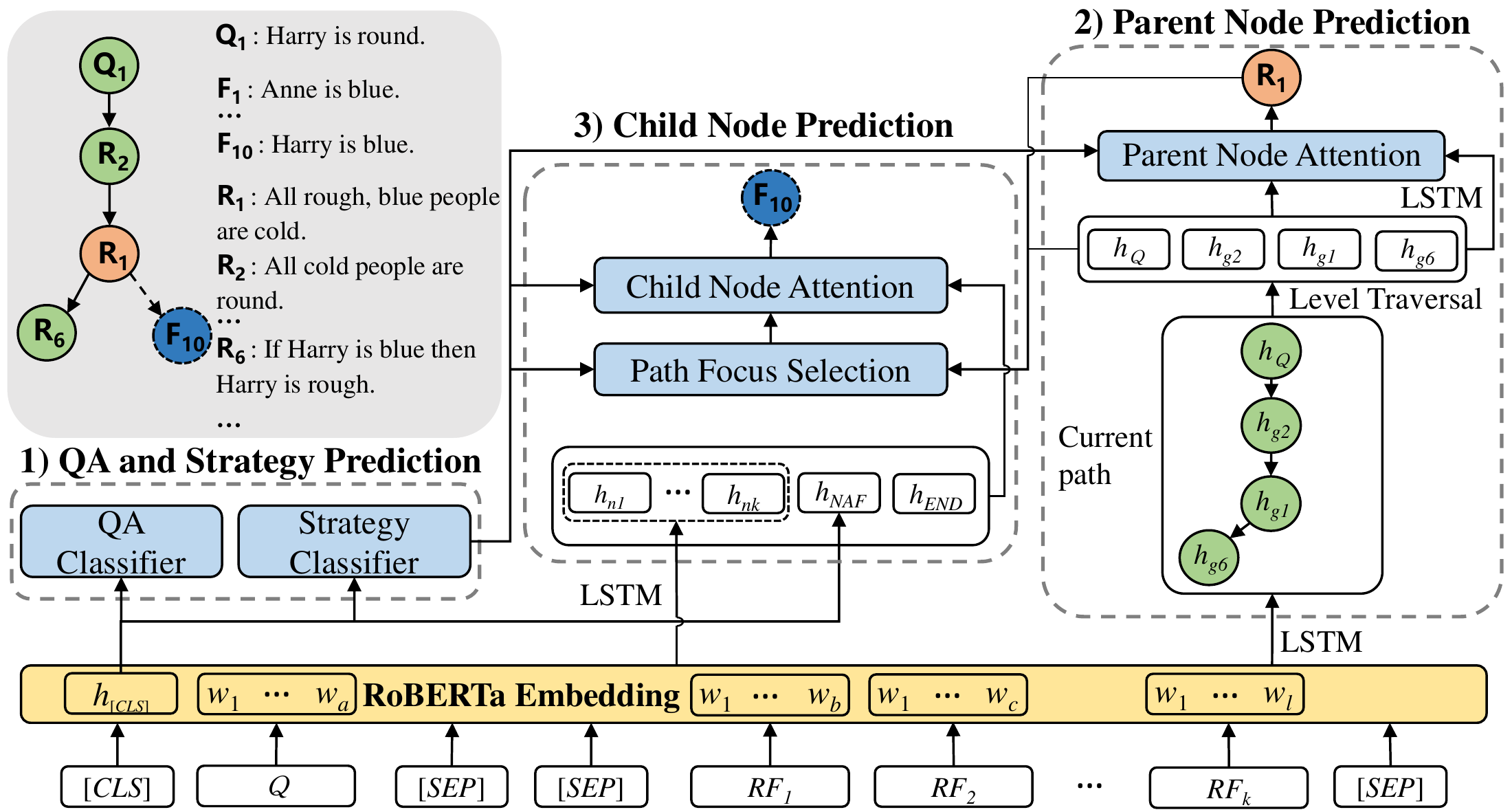}
    \caption{The model architecture of \ourmodel. 1) is only used once at the start, then 2) and 3) are applied iteratively to generate the whole proof. It also illustrates the detailed state when adding F$_{10}$ into the proof (F: facts, R: rules).}
    \label{model}
\end{figure*}

\subsection{Overview}
\label{sec:overview}

The proposed Iterative Backward Reasoning (IBR) model takes $Q$ as the initial node and produce a proof path $P$ backward, from the end node to the start node. Two actions are included at each iteration: (1) \textbf{Predicting the new parent node}, i.e. a node in the derived proof path where a child node will be added (except the first step that only $Q$ exists); (2) \textbf{Predicting the child node}, i.e. the fact or rule in $C$ that will be the child for the selected parent node. After each iteration, a new node and an associated edge are added. After obtaining the whole reasoning path, we remove $Q$ and reverse all edges to get the final proof $P$.

The Figure~\ref{model} illustrates our IBR model, which can be divided into three modules, (1) \textbf{QA and Strategy Prediction}, (2) \textbf{Parent Node Prediction}, and (3) \textbf{Child Node Prediction}. In order to make the question $Q$ can fully interact with context $C$ (facts and rules) and obtain better representations, \ourmodel uses pretrained RoBERTa~\cite{roberta} as the backbone network. The input of RoBERTa is the concatenation of the question $Q$ and the context $C=\{RF_i\}$, separated by special $[SEP]$ token, denoted as  $[CLS] \ Q \ [SEP] \ [SEP] \ C \ [SEP]$.

\ourmodel only uses the QA prediction and strategy prediction modules once at first to predict the answer $A$ and the strategy of the proof (refer to \secref{sec:task_definition}, where the latter one will result in different proof generation procedures.
In order to improve the reasoning efficiency as well as accuracy, instead of using generated intermediate texts~\cite{evr,proofwriter}, all possible nodes (rules and facts) are represented by node embeddings in \ourmodel. The initial state of the proof is only the representation of the question $h_Q$, then the rest of the reasoning path will be constructed based on it. 

Samples with \emph{Fail-Proof} strategy differs from ones with \emph{Proof}, because their proofs are usually short without sub-branches, and only consist of rules due to lacking essential supporting facts. 
To take the advantage of such a property distinction and extend the applicability compared to former models~\cite{evr} that cannot generate proofs for \emph{Fail-Proof} samples, we apply different actions in modules (2) and (3) depending on the output from strategy prediction.

\subsection{QA and Strategy Prediction Module}
\label{sec13}
This module aims to predict the answer $A$ of the question $Q$ and the corresponding strategy $S$ of proof $P$. Since the representation of $[CLS]$ token from pretrained models is proven to have the capability of modeling the whole input, we use it as the input feature for both predictions as they condition the global information. The encoded $[CLS]$ by RoBERTa, $h_{[CLS]}$ is passed to a linear layer and the softmax function $\sigma$ for answer and strategy classification respectively,
\begin{center}
    $ P_{QA} $= $ \sigma $($ f_{QA} $($ h_{[CLS]} $)),
\end{center}

\begin{center}
    $ P_{Strategy} $ = $ \sigma $($ f_{Strategy} $($ h_{[CLS]} $)).
\end{center}
Here, $f_{QA}$ and $f_{Strategy}$ indicate the linear layer for QA classification and strategy classification, respectively. $P_{QA}$ and $P_{Strategy}$ are binary-class probability values, the former one is for values of $A \in \{True, False\}$ while the later one is for values of $S \in \{$\textit{Proof}, \textit{Fail-proof}$\}$.

\subsection{Parent Node Prediction Module}
\label{sec:parent_node_prediction}
This module determines which node in the current reasoning path is going to be the next parent node that a new child node will link to. To better represent the sequential information of each possible node (fact or rule), an LSTM~\cite{hochreiter_lstm} is used to further encode the token-level embedding from RoBERTa. The hidden state in the last step is used as the textual representation $h_{gi}$ of a possible parent node $RF_i$. 

In addition, selecting a node from the existing proof path also needs global and structural modeling on the history path. To make this procedure a more convenient representation that involves the order of reasoning, the path is regarded as a tree structure and nodes are reordered by level traversal from top to down. 
Since $Q$ is always the root node of the tree, e.g., if $Q$ have two children $RF_1$ and $RF_3$, and $RF_1$ has a child $RF_2$, the reordered representation sequence is $[h_Q,h_{g1},h_{g3},h_{g2}]$.
We then utilize another LSTM model to encode the reordered representation sequence of the current reasoning path obtained before, extracting the overall state of the path, which is the hidden state $h_g$ at the last time step in this LSTM.

A parent node attention based on the Transformer attention~\cite{attention} is used to obtain the weights of all possible parents nodes. It takes $h_g$ and the representation sequence of the current path $\mathbf{H}_p=[h_Q,h_{g1} \ldots h_{gt}]$ as input, i.e.
\begin{equation}
\label{eq:att}
    {\rm Att}(h_g, \mathbf{H}_p) = \sigma(f_Q(h_g) (f_K(\mathbf{H}_p))^T / \sqrt{d}),
\end{equation}
where $f_Q$ and $f_K$ indicate linear layers, $\sigma$ is a softmax function, and $d$ is the dimension of $h_g$.
As we discussed in~\secref{sec:overview}, different operations are employed for corresponding strategy types of proofs. 
1) If the predicted proof strategy is \emph{Proof}, we select the node with the highest weight as the parent node $RF_p$. 2) If the predicted proof strategy is \emph{Fail-proof}, we use the last node in the current path, i.e. $h_{gt}$ in $\mathbf{H}_P$, as the parent node $RF_p$, because no sub-branch is included in such proof paths.

\subsection{Child Node Prediction Module}
\label{sec15}
This module decides which node will be added to the proof path and linked to the parent node $RF_p$ we have obtained before.
To derive the representations of candidate child nodes, similar to \secref{sec:parent_node_prediction}, we apply another LSTM model to the encoded RoBERTa embeddings and get $h_{n_{i}}$ for $RF_i$.
Since we discussed a special NAF node in \secref{sec:task_definition} which may contain information from the whole context, we utilize a linear layer $f_{NAF}$ to transform the $[CLS]$ token embedding $h_{[CLS]}$ into its representation $h_{NAF}$. Moreover, we initialize a representation $h_{END}$ for the special END node, indicating that the proof generation process will finish here. 

During selecting the new child node, we need to consider not only the knowledge of the history path, but also the state of the parent node. To better model such relationships, we propose a \textbf{Path Focus Selection} module to generate relevant features before predicting the child node. 
A 2-layer Transformer model along with a LSTM model is introduced. It first encodes the representations of node sequence $\mathbf{H}_p$ from Parent Node Prediction respectively, then fuses their hidden state via a linear layer $f_U$,
\begin{equation}
\label{eq:focus}
    h_F=f_U([{\rm Trans}(h_{gp}, \mathbf{H}_p, \mathbf{H}_p);{\rm LSTM}(\mathbf{H}_p)]).
\end{equation}
Here, $h_{gp}$ is the representation of the selected parent node in \secref{sec:parent_node_prediction}, $f_U$ is the linear layer for feature fusing, while $[\cdot;\cdot]$ stands for concatenation. $q, k, v$ in ${\rm Trans}(q,k,v)$ indicate the inputs corresponding to Query, Key, and Value in a transformer model, and only the hidden state in the last time step is remained in both ${\rm Trans}$ and ${\rm LSTM}$. 
It is worth noting that the LSTM used here is a supplementary knowledge source for a better representation according to our empirical study.
Such an operation results in a feature $h_F$ that is aware of both the history proof path and the parent node that a child will link to.

This feature $h_F$ will then be used in the Child Node Attention to calculate the attention weights on all possible child nodes. Particularly, an attention model same as Eq.~\ref{eq:att} is applied on $h_F$ and a series of child node representations obtained before $\mathbf{H}_c=[h_{n1} \ldots h_{nk}, h_{NAF}, h_{END}]$, and the attention weights are defined as ${\rm Att}(h_F,\mathbf{H}_c)$. It contains all facts and rules in the context, and the special NAF node as well as END node.

Similar to \secref{sec:parent_node_prediction}, we also apply different actions according to our predicted proof strategies before.

\noindent (1) If the strategy is \emph{Proof}, we select the child node with the highest attention weight from all candidates as the new node in the proof path.

\noindent (2) If the strategy is \emph{Fail-proof}, since $RF_p$ is the last node during reasoning and this procedure is a first-order logical under such a situation, there is no need to make complex modeling on the derived path. Therefore, we directly use its parent node representation $h_{gp}$ rather than encoded state from Transformer in Eq.~\ref{eq:focus} to get $h_F$. But LSTM is remained to maintain some basic modeling capability on the path. In child node attention, we mask all fact nodes and select the one with the highest weight among the remaining nodes, because this kind of proof usually only contains rules and such masking can avoid extra errors.

\subsection{Training and Inference}
\label{sec16}
The whole model is trained via binary cross-entropy losses from all three above modules jointly,
\begin{center}
    $L = L_{QA} + L_{Parent} + L_{Child} + \alpha * L_{Strategy}$.
\end{center}
$ L_{QA} $ and $ L_{Strategy}$ correspond to the loss of QA prediction and strategy prediction, respectively. $\alpha$ is a hyperparameter to reweigh the influence of [\textit{CLS}] token. $L_{Parent}$ is the loss for parent node prediction, where the cross-entropy is calculated between the attention weight vector and a one-hot vector indicating the gold parent node.
$L_{Child}$ is in a similar way on child node prediction. Note that samples labeled as \emph{Fail-proof} strategy are not involved in the training of parent node prediction. As all their proof paths are chains and the new parent node is always the last node added to the path, so learning about these data may introduce model bias. To determine the gold reasoning order used as the target for training, we set a higher priority of fact nodes than rule nodes, as the clearer subject information is involved in facts. E.g., for a parent node with multiple children, the gold reasoning order of child node prediction is NAF nodes first, then fact nodes, and finally rule nodes. If there are more than one fact or rule nodes, \ourmodel randomly swaps their order within each type at different training epochs.

During inference, \ourmodel first makes predictions on the answer $A$ and strategy $S$, then generate the parent node and child node iteratively, until the special END node is predicted as the new child node. IBR uses beam search to keep the top-K best proof paths at each proof generation step and select the best one as the final prediction, where the beam size is set as 8.



\section{Experiments}
\label{sec:experiment}

Following former studies~\citep{prover,probr}, we evaluate our \ourmodel\footnote{Refer to Appendix~\ref{app:implementation_details} for implementation details.} on three datasets and four settings including fully-supervised training, training using fewer samples, testing on out-of-domain samples, and generalization to more complex proofs or language.


\subsection{Setup}
\label{sec:exp_setup}
\paragraph{Datasets.}
Experiments are conducted on three datasets raised by \citet{clark_synthetic_dataset}\footnote{More details are given in Appendix~\ref{app:dataset_details}}, where we use the same test split as previous works for fair comparison:
\begin{itemize}[wide=0\parindent, noitemsep, topsep=0pt]
    \item \textbf{DU0-DU5}: Five synthesized datasets created by translating hand-crafted rules and formal language to natural language. It is divided by the highest depth of proof, where DU stands for "Depth Upto" (DU=0,1,2,3,5). Data in higher DU values also contain samples with lower depth. Note that proofs in DU0 only have one supporting or opposing fact. All related results are reported on DU5 test split.
    \item \textbf{Bird-Electricity}: It is a test-only dataset that contains samples about birds and electric circuits. It is generated in the same way as DU0-DU5, but is in different domains from DU0-DU5.
    \item \textbf{ParaRules}: This dataset consists of 40k questions expressed in paraphrased natural language based on synthetic data, which is created by crowdsourcing. Multiple facts get together in one statement here rather than separated in DU0-DU5.
\end{itemize}

\paragraph{Baselines.} We consider the following baselines\footnote{Results of baselines are obtained from the original papers or by running the released code.}.
\begin{itemize}[wide=0\parindent, noitemsep, topsep=0pt]
    \item \textbf{RuleTaker (RT)}~\cite{clark_synthetic_dataset}: a RoBERTa based model that only predicts answers.
    \item \textbf{PROVER (PV)}~\cite{prover}: a method that treats the proof as a graph and predicts all its nodes and edges at once, also using RoBERTa model as the backbone, same as \ourmodel.
    \item \textbf{PROBR (PB)}~\cite{probr}: it improves PROVER by introducing the probabilistic graph that jointly considers the answer, nodes and edges.
    \item \textbf{EVR}~\cite{evr}: an iterative model that predicts the next proof item by generating textual sub-questions based on logical operator. Note that this model is \textbf{not applicable} for samples whose \textbf{proof strategy is \emph{Fail-proof}} discussed in \secref{sec:task_definition}, so we make comparison with it separately.
\end{itemize}

\paragraph{Metrics.} 
We closely follow previous works to evaluate the performance of models via answer prediction (QA) accuracy and proof generation (PA) accuracy.
Since some samples may have multiple gold proofs, a generated proof will be considered correct, as long as its nodes and edges match with the nodes and the edges in any of the gold proofs. Full Accuracy (FA) is also included, where a sample is regarded as correct only both the predicted answer and proof are correct. 

\subsection{Results under Fully-Supervised Training}
\label{sec:main_results}
We train \ourmodel on the training split of the DU5 dataset and evaluate on the test split of DU5. We compare the performance of IBR with baselines except for EVR in Table~\ref{tab:main_results}, while with EVR in Table~\ref{tab:main_results_evr} where only partial test split is included, excluding samples whose proof strategy is \emph{Fail-proof}. Because EVR always fails on these samples (EVR on these excluded samples is given in Appendix~\ref{app:evr_fail_results}).

Obviously, \ourmodel achieves the best proof generation accuracy (PA) as well as full accuracy (FA) among all baseline models, on samples with every depth. Our model also shows a greater advantage on samples with deeper proof path, e.g., 81.7 vs. 72.2 on PA when depth is 5, illustrating the superiority of iterative models on complex proof paths. Besides, despite not being the best in answer accuracy (QA), there is a very narrow gap between our model and the best one, which proves that \ourmodel is still a comprehensive model covering both subtasks. When compared to EVR, also an iterative model, \ourmodel shows significantly stronger performance on all metrics, benefiting from our elaborate two-fold reasoning process at each step.


\begin{table}[]
\setlength{\belowcaptionskip}{-0.2cm}
\centering
\small
\setlength{\tabcolsep}{3.2pt}
\renewcommand{\arraystretch}{1}
\begin{tabular}{llrrrrrrr}
\toprule
& \textbf{D}   & \textbf{0}   & \textbf{1}    & \textbf{2}    & \textbf{3}   & \textbf{4}   & \textbf{5}   & \textbf{all} \\
\midrule 
& \textbf{Cnt} & 6299         & 4434          & 2915          & 2396         & 2134         & 2003         & 20192 \\
\midrule
\multirow{4}{*}{\textbf{QA}}
& RT &           100          & 98.4          & 98.4          & 98.8         & 99.2         & 99.8         & 99.2 \\
& PV &           100          & 99.0          & 98.8          & 99.1         & 98.8         & 99.3         & 99.3 \\
& PB &           100          & \textbf{99.9} & \textbf{99.9} & \textbf{100} & \textbf{100} & \textbf{100} & \textbf{99.9} \\
& \textbf{IBR} & \textbf{100} & 99.2          & 99.2          & 98.9         & 99.3         & 99.6         & 99.4 \\
\midrule
\multirow{3}{*}{\textbf{PA}}
& PV &           98.4          & 93.2          & 84.8          & 80.5         & 72.5         & 65.1         & 87.1 \\
& PB &           98.4          & 94.3          & 86.1          & 82.0         & 76.1         & 72.2         & 88.8 \\
& \textbf{IBR} & \textbf{99.5} & \textbf{95.6} & \textbf{93.0} & \textbf{90.7}& \textbf{86.5}& \textbf{81.7}& \textbf{93.5} \\
\midrule
\multirow{3}{*}{\textbf{FA}}
& PV &           98.4          & 93.1          & 84.8          & 80.5         & 72.4         & 65.1         & 87.1 \\
& PB &           98.4          & 94.3          & 86.1          & 82.0         & 76.1         & 72.2         & 88.8 \\
& \textbf{IBR} & \textbf{99.5} & \textbf{95.6} & \textbf{92.9} & \textbf{90.7}& \textbf{86.5}& \textbf{81.6}& \textbf{93.5} \\
\bottomrule
\end{tabular}
\caption{Results of different models on varying proof depth (\textbf{D}) under the fully-supervised setting. Cnt: sample count, RT: RuleTaker, PV: PROVER, PB: PROBR.}
\label{tab:main_results}
\end{table}

\begin{table}
\setlength{\belowcaptionskip}{-0.2cm}
\small
\centering
\setlength{\tabcolsep}{3.2pt}

\begin{tabular}{llrrrrrrr}
\toprule
& \text{D} & \textbf{0} & \textbf{1} & \textbf{2} & \textbf{3} & \textbf{4} & \textbf{5} & \textbf{all}\\
\midrule
& \textbf{Cnt} & 1934         & 1934          & 1934          & 1934         & 1934         & 1934         & 11604 \\
\midrule
\multirow{2}{*}{\textbf{QA}}
& EVR &           99.4          & 99.3 & 96.9 & 93.3 & 88.9 & 88.3 & 94.4 \\
& \textbf{IBR} & \textbf{100} & \textbf{99.3}          & \textbf{99.6}          & \textbf{99.3}         & \textbf{99.6}         & \textbf{99.5}         & \textbf{99.5} \\
\midrule
\multirow{2}{*}{\textbf{PA}}
& EVR &           95.8          & 92.5          & 87.7          & 79.3         & 77.3         & 68.8         & 83.6 \\
& \textbf{IBR} & \textbf{98.8} & \textbf{96.4} & \textbf{94.7} & \textbf{92.2}& \textbf{88.7}& \textbf{83.6}& \textbf{92.4} \\
\midrule
\multirow{2}{*}{\textbf{FA}}
& EVR &           95.8          & 92.5          & 87.7          & 79.3         & 77.3         & 68.8         & 83.6 \\
& \textbf{IBR} & \textbf{98.8} & \textbf{96.3} & \textbf{94.6} & \textbf{92.2}& \textbf{88.7}& \textbf{83.5}& \textbf{92.3} \\
\bottomrule

\end{tabular}
\caption{Results of IBR and EVR on a partial test split of DU5 (exclude \emph{Fail-proof} samples). The models are trained on the training split of DU5.}
\label{tab:main_results_evr}
\end{table}

\subsection{Using Fewer Training Samples}
\label{sec11}
We also explore the performance of \ourmodel when training using fewer data, ranging from 10k to 30k to all the examples (70k) in DU5. 
The comparison between our model, PROVER (PV), and PROBR (PB) is shown in Table~\ref{tab:few_shot}, in all three metrics. 
Our model significantly has the best proof generation performance than the other two baselines in all cases, due to the iterative architecture requiring less global modeling capability and thus fewer training samples. Although PB shows a promising answer prediction accuracy under fewer-data settings, the performance of \ourmodel is close to it while better than PV, e.g., 94.3 vs. 87.1 under 10k. In addition, in Table~\ref{tab:few_shot_evr}, we also compare with EVR under the same settings but using a different test set that excludes \emph{Fail-proof} samples. EVR outperforms \ourmodel under the 10k setting for proof generation, but \ourmodel is stronger if more training samples are available.

\begin{table}[t!]
\setlength{\belowcaptionskip}{-0.2cm}
\small
\centering
\setlength{\tabcolsep}{2.5pt}
\renewcommand{\arraystretch}{1}{
\begin{tabular}{lccccccccc}
\bottomrule
\multirow{2}{*}{Data} & \multicolumn{3}{c}{\textbf{QA}}      & \multicolumn{3}{c}{\textbf{PA}}                        & \multicolumn{3}{c}{\textbf{FA}}               \\ 
\specialrule{0pt}{0pt}{2pt}
\cline{2-10} 
\specialrule{0pt}{2pt}{0pt}
{}           & PV   & PB            & \textbf{IBR} & PV            & PB            & \textbf{IBR}          & PV   & PB            & \textbf{IBR}          \\ \midrule
70k      & 99.3 & \textbf{99.9} & 99.4 & 87.1          & 88.8          & \textbf{93.5} & 87.1 & 88.8          & \textbf{93.5} \\
              30k            & 97.8 & \textbf{99.9} & 98.3 & 72.5          & 86.8          & \textbf{89.8} & 72.4 & 86.8          & \textbf{89.7} \\
            10k            & 87.1 & \textbf{99.9} & 94.3 & 44.0          & 72.4          & \textbf{75.7} & 42.7 & 72.3          & \textbf{75.4}\\ \bottomrule
\end{tabular}}
\caption{Performance comparison using fewer training samples among IBR, PROVER (PV), and PROBR (PB) on the full test split of DU5 after trained on partial DU5 samples.}
\label{tab:few_shot}
\end{table}

\begin{table}[t!]
\setlength{\belowcaptionskip}{-0.2cm}
\small
\centering
\setlength{\tabcolsep}{4pt}
\renewcommand{\arraystretch}{1}{
\begin{tabular}{lcccccc}
\bottomrule
\multirow{2}{*}{Data} & \multicolumn{2}{c}{\textbf{QA}}      & \multicolumn{2}{c}{\textbf{PA}}                        & \multicolumn{2}{c}{\textbf{FA}}               \\ 
\specialrule{0pt}{0pt}{2pt}
\cline{2-7} 
\specialrule{0pt}{2pt}{0pt}
{}           & EVR   & \textbf{IBR} & EVR  & \textbf{IBR}          & EVR   & \textbf{IBR}          \\ \midrule
70k      & 94.4 & \textbf{99.5} & 83.6          & \textbf{92.4} & 83.6  & \textbf{92.3} \\
              30k            & 95.7 & \textbf{99.4} & 84.4          & \textbf{88.2} & 84.4 & \textbf{88.1} \\
            10k            & 96.2 & \textbf{97.9} & \textbf{82.8}          & 71.2 & \textbf{82.8} & 70.8\\ \bottomrule
\end{tabular}}
\caption{Performance comparison using fewer training samples among EVR and IBR on partial test split of DU5 (without \emph{Fail-proof} samples) after trained on partial DU5 samples.}
\label{tab:few_shot_evr}
\end{table}

\subsection{Evaluation of Out-of-Domain Data}
\label{sec:zero_shot}
We further test the out-of-domain performance of \ourmodel against baselines on Birds-Electricity dataset to evaluate their robustness, where B1 and B2 are two sets from the birds domain, and E1-E4 are four sets from the electricity domain.
Results are shown in Table~\ref{tab:zero_shot} and Table~\ref{tab:zero_shot_evr}. Note that \emph{Fail-proof} samples are still not involved in the comparison for EVR. 
Overall, our \ourmodel achieves 2.5\% promotion in PA while an equivalent result on QA, compared to PROVER. Despite being the best one on QA, PROBR is also defeated by \ourmodel on both PA and FA. In addition, our model shows more improvement on the hardest E3 and E4 subsets, which further verifies its robustness. When it comes to EVR, we can find its cross-domain capability is relatively weak as it sees a significant drop in PA, and \ourmodel is superior to it without any doubt. Because the cross-domain generation for intermediate texts is much harder, our usage of high-level node features to finished reasoning can alleviate this challenge.

\begin{table}[t!]
\setlength{\belowcaptionskip}{-0.2cm}
\small
\centering
\setlength{\tabcolsep}{3pt}
\renewcommand{\arraystretch}{1}{
\begin{tabular}{llrrrrrrr}
\toprule
& \textbf{Test}   & \textbf{B1}   & \textbf{B2}    & \textbf{E1}    & \textbf{E2}   & \textbf{E3}   & \textbf{E4}   & \textbf{all} \\
\midrule 
& \textbf{Cnt} & 40         & 40          & 162          & 180         & 624         & 4224         & 5270 \\
\midrule
\multirow{4}{*}{\textbf{QA}}
& RT &           97.5          & 100.0          & 96.9          & 98.3         & 91.8         & 76.7         & 80.1 \\
& PV &           95.0          & 95.0          & 100.0          & 100.0         & 89.7         & 84.8         & 86.5 \\
& PB &           100.0          & \textbf{100.0} & 100.0 & 100.0 & \textbf{98.2} & \textbf{95.6} & \textbf{96.3} \\
& \textbf{IBR} & \textbf{100.0} & 97.5          & \textbf{100.0}          & \textbf{100.0}         & 89.2         & 84.1         & 86.0 \\
\midrule
\multirow{3}{*}{\textbf{PA}}
& PV &           92.5          & 95.0          & 95.1          & 91.7         & 72.3         & 80.6         & 80.7 \\
& PB &           100.0          & 100.0          & \textbf{97.5}          & 93.3         & 79.3         & 77.7         & 79.3 \\
& \textbf{IBR} & \textbf{100.0} & \textbf{100.0} & 95.6 & \textbf{94.4}& \textbf{80.2}& \textbf{82.4}& \textbf{83.2} \\
\midrule
\multirow{3}{*}{\textbf{FA}}
& PV &           92.5          & 95.0          & 95.1          & 91.7         & 71.8         & 80.6         & 80.5 \\
& PB &           100.0          & \textbf{100.0}          & \textbf{97.5}          & 93.3         & \textbf{79.3}         & 77.7         & 79.3 \\
& \textbf{IBR} & \textbf{100.0} & 97.5 & 95.6 & \textbf{94.4}& 78.2& \textbf{82.4}& \textbf{82.9} \\
\bottomrule
\end{tabular}}
\caption{Out-of-domain performance comparison among RuleTakers (RT), PROVER (PV), and PROBR (PB) on Birds-Electricity dataset after training on DU5.}
\label{tab:zero_shot}
\end{table}

\begin{table}[t!]
\setlength{\belowcaptionskip}{-0.2cm}
\small
\centering
\setlength{\tabcolsep}{3pt}
\renewcommand{\arraystretch}{1}{
\begin{tabular}{llrrrrrrr}
\toprule
& \textbf{Test}   & \textbf{B1}   & \textbf{B2}    & \textbf{E1}    & \textbf{E2}   & \textbf{E3}   & \textbf{E4}   & \textbf{all} \\
\midrule 
& \textbf{Cnt} & 28         & 28          & 72          & 90         & 312         & 1206         & 1736 \\
\midrule
\multirow{2}{*}{\textbf{QA}}
& EVR &           67.8          & 64.2          & 83.3          & 80.0         & 76.2         & 83.8         & 81.6 \\
& \textbf{IBR} & \textbf{100.0} & \textbf{96.4}          & \textbf{100.0}          & \textbf{100.0}         & \textbf{92.9}         & \textbf{100.0}         & \textbf{98.6} \\
\midrule
\multirow{2}{*}{\textbf{PA}}
& EVR &           32.1          & 35.7          & 58.3          & 50.0         & 45.5         & 70.3         & 63.1 \\
& \textbf{IBR} & \textbf{100.0} & \textbf{100.0} & \textbf{91.6} & \textbf{91.1}& \textbf{91.3}& \textbf{95.2}& \textbf{94.3} \\
\midrule
\multirow{2}{*}{\textbf{FA}}
& EVR &           32.1          & 32.1          & 58.3          & 50.0         & 45.5         & 70.3         & 63.1 \\
& \textbf{IBR} & \textbf{100.0} & \textbf{96.4} & \textbf{91.6} & \textbf{91.1}& \textbf{87.1}& \textbf{95.2}& \textbf{93.5} \\
\bottomrule
\end{tabular}}
\caption{Out-of-domain performance comparison among EVR and IBR on partial Birds-Electricity dataset (exclude \emph{Fail-proof} samples) after training on DU5.}
\label{tab:zero_shot_evr}
\end{table}

\subsection{Generalization Ability}
\label{sec:generalization}
\textbf{Generalize to higher depths.} Following the previous work~\citep{probr}, we test the generalization ability of \ourmodel by first training the model on the training splits of DU0, DU1, DU2, and DU3, 
then test them on the test split of DU5 with deeper proof paths respectively\footnote{We remove the position embedding in path focus selection to proceed to this test, see Appedix~\ref{app:implementation_details} for details}.
Results are shown in Table~\ref{tab:generalization_depth}.
We notice that all models suffer performance degeneration especially when the proof depth of the training set is lower, because it is hard for the model to learn complex reasoning based on simple proof paths. However, \ourmodel still realizes the best performance in terms of PA and FA, especially on \textbf{DU3}, where it gets 4.2\% PA/FA promotion to PROBR and even outperforms PROVER trained on the whole \textbf{DU5} data. These observations again prove that iterative approaches can better learn the detailed reasoning step by step, obtaining a better generalization capability than at-once models.

\begin{table}[t!]
\setlength{\belowcaptionskip}{-0.2cm}
\small
\centering
\setlength{\tabcolsep}{3pt}
\renewcommand{\arraystretch}{1}{
\begin{tabular}{lrrrrrrrrr}
\bottomrule
\multirow{2}{*}{Data} & \multicolumn{3}{c}{\textbf{QA}}                      & \multicolumn{3}{c}{\textbf{PA}}               & \multicolumn{3}{c}{\textbf{FA}}      \\ 
\specialrule{0pt}{0pt}{2pt}
\cline{2-10} 
\specialrule{0pt}{2pt}{0pt} 
 & PV            & PB            & \textbf{IBR} & PV            & PB            & \textbf{IBR} & PV   & PB            & \textbf{IBR} \\ \midrule
\textbf{DU0}                                                             & \textbf{68.7} & 56.9          & 53.5          & 44.4          & \textbf{50.7} & 47.0          & 42.8 & 41.3          & \textbf{47.0} \\
\textbf{DU1}                                                             & 73.7          & \textbf{97.7} & 73.1          & 63.8 & 63.9          & \textbf{64.6}          & 61.9 & 63.9 & \textbf{64.5}          \\
\textbf{DU2}                                                             & 89.6          & \textbf{99.9} & 89.6          & 72.6          & 74.5 & \textbf{76.3}          & 72.3 & 74.4 & \textbf{76.2}          \\
\textbf{DU3}                                                             & 98.6          & \textbf{99.9} & 98.6          & 79.1          & 83.2          & \textbf{87.4} & 79.1 & 83.2          & \textbf{87.4} \\ \midrule
\textbf{DU5}                                                             & 99.3          & \textbf{99.9} & 99.4          & 87.1          & 88.8          & \textbf{93.5} & 87.1 & 88.8          & \textbf{93.4} \\ \bottomrule
\end{tabular}}
\caption{Performance of generalization ability between PROVER (PV), PROBR (PB), and \ourmodel when testing on the test split of DU5, after trained on DU0, DU1, DU2, DU3, and DU5, respectively.}
\label{tab:generalization_depth}
\end{table}

\begin{table}[t!]
\setlength{\belowcaptionskip}{-0.2cm}
\small
\centering
\setlength{\tabcolsep}{5pt}
\begin{tabular}{llrrrrrr}
\toprule
& \textbf{D}   & \textbf{0}   & \textbf{1}    & \textbf{2}    & \textbf{3}   & \textbf{4}     & \textbf{all} \\
\midrule 
& \textbf{Cnt} & 2968         & 2406          & 1443          & 1036         & 142         & 8008\\
\midrule
\multirow{3}{*}{\textbf{QA}}
& PV &           99.7          & 98.6          & 98.2          & 96.5         & 88.0         & 98.4\\
& PB &           99.8          & \textbf{99.7} & \textbf{99.9} & \textbf{99.8} & \textbf{100} & \textbf{99.8}\\
& \textbf{IBR} & \textbf{99.9} & 98.8          & 97.5          & 96.3         & 88.7         & 98.4\\
\midrule
\multirow{3}{*}{\textbf{PA}}
& PV &           99.5          & 98.0          & 88.9          & 90.0         & 76.1         & 95.4\\
& PB &           99.5          & 98.0          & 88.9          & \textbf{90.1}         & \textbf{82.4}         & 95.6\\
& \textbf{IBR} & \textbf{99.8} & \textbf{98.8} & \textbf{91.1} & 89.0 & 75.3& \textbf{95.9}\\
\midrule
\multirow{3}{*}{\textbf{FA}}
& PV &           99.4          & 97.3          & 88.7          & 89.9         & 76.1         & 95.1\\
& PB &           99.4          & 98.0          & 88.9          & \textbf{90.1}         & \textbf{82.4}         & 95.5\\
& \textbf{IBR} & \textbf{99.7} & \textbf{98.1} & \textbf{90.9} & 89.0 & 75.3 & \textbf{95.7}\\
\bottomrule
\end{tabular}
\caption{Performance on ParaRules test set, after trained on combined D3+ParaRules training partitions, including PROVER (PV), PROBR (PB), and \ourmodel.}
\label{tab:para_results}
\end{table}

\paragraph{Generalize to complex language.}
We also evaluate whether \ourmodel can be applied to samples where questions and statements are expressed in more human-like natural language. Following \citet{clark_synthetic_dataset}, we train models on the combined training partitions of DU3 and ParaRules then test them on the ParaRules test set. To our best knowledge, it is the dataset that is closest to real-world applications. Table~\ref{tab:para_results} demonstrates that our model sees a slight promotion in PA/FA while a similar accuracy as PROVER in QA, indicating that \ourmodel still has good applicability when doing reasoning on more complicated and natural texts.

\section{Analysis}

\subsection{Ablation Study}
\label{sec:ablation_study}

To explore the effects between different components in our model, we consider the following ablations: 1) \ourmodel+Gold-Parent: given the gold parent nodes during inference to explore the accuracy of child node prediction; 2) \ourmodel+Gold-Child: given the gold child nodes to verify the accuracy of parent node prediction; 3) \textit{w/o} QA: removing QA task in loss to check its impact on proof generation; 4) \textit{w/o} node LSTM: using mean pooling rather than LSTM encoding to get the representations of nodes; 5) \textit{w/o} focus LSTM: Removing the supplementary LSTM in path focus selection. 

Results on the whole DU5 test split are given in Table~\ref{tab:ablation}. As the numeric performance shows, giving either gold parent nodes or gold child nodes can benefit the performance especially the later one. This signifies that our parent node prediction achieves promising accuracy while the prediction of child nodes can be further improved. Moreover, \ourmodel can still learn to generate proofs without supervision from answers. And LSTM encoders attribute to a better representation of both the nodes and the path that has been derived. 

\begin{table}[t!]
\setlength{\belowcaptionskip}{-0.2cm}
\small
\centering
\renewcommand{\arraystretch}{1}
\begin{tabular}{lrrr}
\bottomrule
\textbf{Models} & QA & PA & FA \\
\midrule
\ourmodel & 99.4                   & 93.5                   & 93.5                   \\
\quad \ourmodel+Gold-Parent     & 99.4                   & 95.6                   & 95.3                   \\
\quad \ourmodel+Gold-Child     & 99.4                   & 99.6                   & 99.3                   \\
\quad \textit{w/o} QA         &\multicolumn{1}{c}{-}                        &93.7                        &\multicolumn{1}{c}{-}                        \\
\quad \textit{w/o} node LSTM            & 99.5                   & 93.2                   & 93.2                   \\
\quad \textit{w/o} focus LSTM & 99.6                   & 92.6                   & 92.4                   \\
\bottomrule
\end{tabular}
\caption{Results of ablation studies on DU5 dataset. We use \ourmodel as the backbone.}
\label{tab:ablation}
\end{table}

\subsection{Latency Analysis}
\label{sec11}
\begin{figure}[t!]
\setlength{\abovecaptionskip}{0.1cm}
\setlength{\belowcaptionskip}{-0.0cm}
    \centering
    \includegraphics[width=0.9\linewidth]{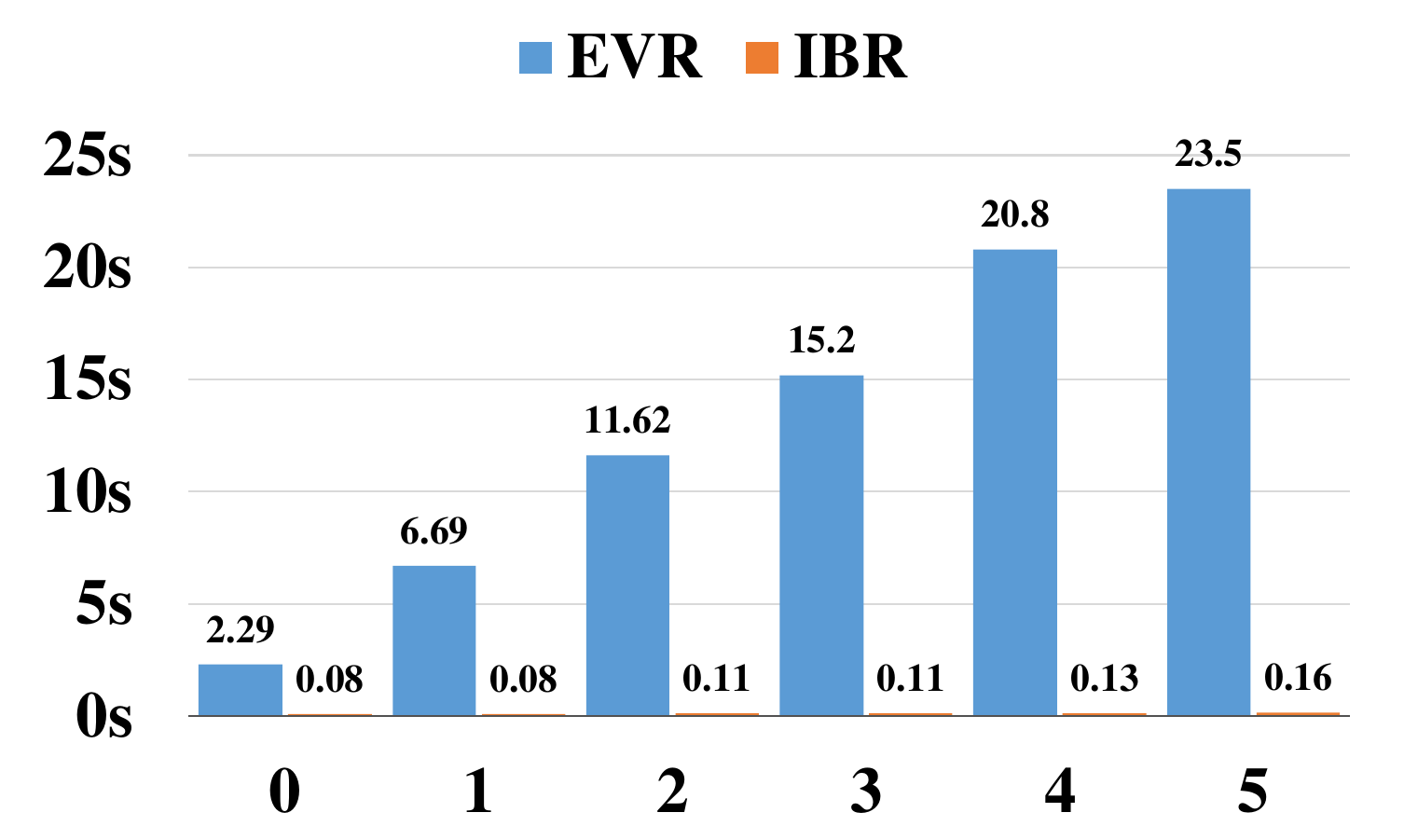}
    \caption{Per-sample inference runtime (in second) of EVR and IBR on DU5 dataset with varying depths.}
    \label{fig:runtime}
\end{figure}

\begin{figure}[t!]
\setlength{\abovecaptionskip}{0.1cm}
\setlength{\belowcaptionskip}{-0.0cm}
    \centering
    \includegraphics[width=0.9\linewidth]{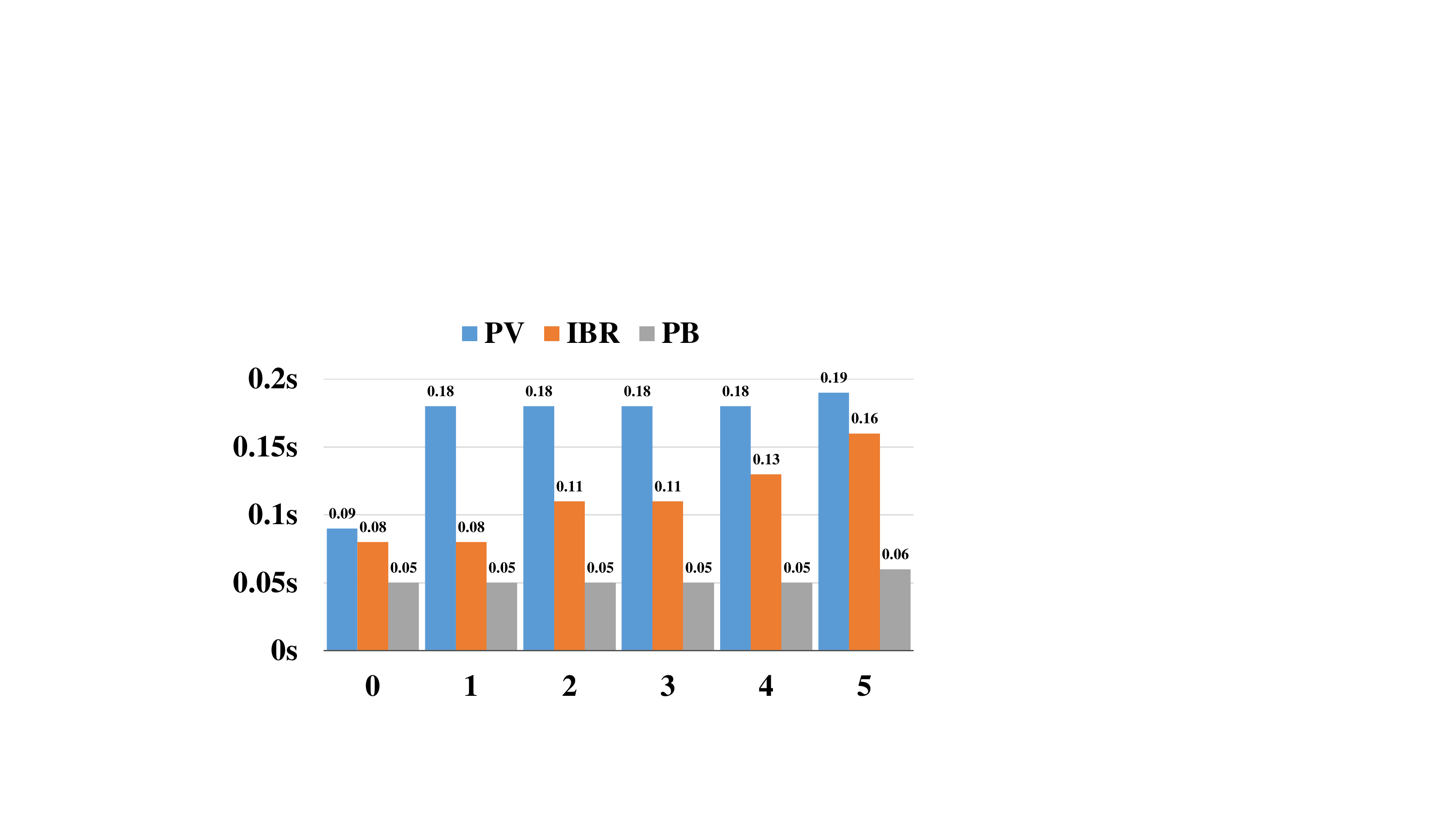}
    \caption{Per-sample inference runtime (in second) of PROVER (PV), IBR, and PROBR (PB) on DU5 dataset with varying depths.}
    \label{fig:runtime_PV_PB}
\end{figure}

To demonstrate the computational efficiency of IBR, we compare the per sample inference time of \ourmodel with EVR, also an iterative proof generation model, on the test split of DU5. Additionally, we also compare the per sample inference time of \ourmodel with PROVER and PROBR, both at-once models. All models are tested on one \texttt{NVIDIA Tesla-V100} GPU with the same batch size and the beam size of IBR sets to 1 for a fair comparison. 
As shown in Figure~\ref{fig:runtime}, our IBR could achieve up to $\times$119.5 speedup compared with EVR, benefiting from our reasoning based on node and path features rather than intermediate texts. It is also noticeable that the runtime of EVR grows linearly with depth, while such an effect is slight on our model. Because EVR needs to infer on all contexts at every step, but \ourmodel uses a simplified parent node prediction based on the derived path. Figure~\ref{fig:runtime_PV_PB} illustrates that IBR is also faster than PROVER because PROVER has some constraints during post-processing in inference, like ensuring proof connectivity, which takes extra time.

\section{Conclusion}
This paper presents \ourmodel, a proof generation model via iterative backward reasoning for rule-based QA tasks. 
We equip the reasoning procedure with detailed hidden state tracking by predicting nodes and edges in the proof path iteratively backward from the question, and allow the model to reason on the elaborate representations of nodes and history paths. Our model is more interpretable than previous at-once models, and is also more effective and efficient than former iterative models. Experiments also demonstrate the superiority of \ourmodel to various baselines on proof generation under various settings.

\section*{Acknowledgements}
This work was partially supported by the National Natural Science Foundation of China (61876053, 62006062, 62176076), the Shenzhen Foundational Research Funding  JCYJ20210324115614039, Shenzhen Science and Technology Program JSGG20210802154400001, and the Joint Lab of HITSZ and China Merchants Securities.




\bibliography{anthology,custom}
\bibliographystyle{acl_natbib}

\newpage\newpage
\appendix

\section{Appendix}

\subsection{Implementation Details}
\label{app:implementation_details}

\begin{table}[h]
\setlength{\abovecaptionskip}{0.1cm}
\setlength{\belowcaptionskip}{-0.0cm}
\centering
\begin{tabular}{lc}
\toprule
\textbf{Parameter}                & \textbf{Value} \\ 
\midrule
Training Epochs                   & 8              \\
Optimizer                         & AdamW          \\
Batch Size                        & 16             \\
RoBERTa Learning rate             & 1e-5           \\
QA and Strategy Pre Learning rate & 1e-5           \\
Parent Node Pre Learning rate     & 2e-4           \\
Child Node Pre Learning rate      & 5e-4           \\
All LSTM Learning rate            & 1e-3           \\
Dropout Rate                      & 0.1            \\ 
LSTM hidden state for parent node & \multirow{2}{*}{1024} \\
and child node encoding           &                \\
LSTM hidden state for path encoding& \multirow{2}{*}{1024} \\
in parent node prediction          &                \\
Transformer hidden state in path  & \multirow{2}{*}{1024} \\
focus selection                   &                 \\
LSTM hidden state in path focus   & \multirow{2}{*}{256} \\
selection                   &                 \\
Seed                      & 42            \\ 
\bottomrule
\end{tabular}
\caption{Implementation details of IBR.}
\label{tab:implementation_details}
\end{table}
We implement our model based on PyTorch along with Huggingface-Transformers toolkit\footnote{\url{https://github.com/huggingface/transformers}}. We use RoBERTa$_{\rm Large}$ model\footnote{\url{https://huggingface.co/roberta-large}} as our backbone encoder to generate token-level representations. 
Table~\ref{tab:implementation_details} shows the implementation details of IBR, including learning rates for different modules.
All linear layers used in our model have one layer.
The model trained after 8 epochs will be used in the evaluation.
We remove functional words without lexical meaning like "a" and "the" from facts, rules, and questions to shorten the input length, so each training epoch takes about 2 hours. We select these hyper-parameters according to tuning them empirically based on the performance.
All experiments are run on NVIDIA Tesla-V100 GPUs. The main experiment performance of IBR fluctuates by one point. 

\subsection{Dataset Details}
\label{app:dataset_details}

\begin{table}[t!]
\setlength{\abovecaptionskip}{0.1cm}
\setlength{\belowcaptionskip}{-0.0cm}
\centering
\small
\setlength{\tabcolsep}{2.0pt}
\renewcommand{\arraystretch}{1}
\begin{tabular}{llcccc}
\toprule
Split & D & Num & \emph{Fail-proof} Num & \emph{Proof} Num & Avg. Node \\
\midrule
\multirow{7}{*}{\textbf{Train}} & 0 & 21,359 & 14,597 & 6,762 & 0.62 \\
& 1 & 15,380 & 8,618 & 6,762 & 1.82 \\
& 2 & 10,112 & 3,350 & 6,762 & 3.37 \\
& 3 & 8,389 & 1,627 & 6,762 & 4.98 \\
& 4 & 7,456 & 694 & 6,762 & 6.90 \\
& 5 & 6,987 & 225 & 6,762 & 9.26 \\
& all & 69,683 & 29,111 & 40,572 & 3.35 \\
\bottomrule
\multirow{7}{*}{\textbf{Test}} & 0 & 6,299 & 4,365 & 1,934 & 0.59 \\
& 1 & 4,434 & 2,500 & 1,934 & 1.77 \\
& 2 & 2,915 & 981 & 1,934 & 3.36 \\
& 3 & 2,396 & 462 & 1,934 & 4.99 \\
& 4 & 2,134 & 200 & 1,934 & 6.98 \\
& 5 & 2,003 & 69 & 1,934 & 9.47 \\
& all & 20,181 & 8,577 & 11,604 & 3.33 \\
\bottomrule
\end{tabular}
\caption{The statistics of train and test split in DU5 dataset. \emph{Fail-proof} and \emph{Proof} indicate different proof strategies we discussed in \secref{sec:task_definition}. Avg. Node indicates the average node number in a proof path. }
\label{tab:du_statistics}
\end{table}

We next introduce the details of the three datasets used in our experiment. All of them are firstly applied in rule-based QA and proof generation tasks in \citealp{clark_synthetic_dataset}. 

\paragraph{DU0-DU5:} A series of synthesized datasets where rules and facts are all generated via manually designed logical programming, while questions are generated by combining random logical operations among them. Data are divided into 5 subsets according to their maximum reasoning depth (D) in the proof path, D = 0, 1, 2, 3, 5. There are 100k questions in each subset, where 70k / 10k / 20k samples in the train / validation / test partition respectively. D = 0 means that the question can be proven directly using a fact in contexts. In our experiment in \secref{sec:experiment}, we only use the data from DU5 for testing because it covers all possible depths, while the train set is the train split in DU5 except \secref{sec:generalization}, where we use train split from DU0, DU1, DU2 and DU3 for training. We provide some statistics of DU5 in Table~\ref{tab:du_statistics}.

\paragraph{Birds-Electricity:} It is a set of data that only contains 5k test samples for the evaluation of robustness and out-of-domain performance of models. The Birds data only require reasoning up to depth 1 and 2 (B1 and B2), while Electricity data have reasoning depths ranging from 1 to 4. Both of them include new vocabulary that is not included in DU0-DU5.

\paragraph{ParaRules:} A more challenging dataset contains paraphrased samples on the synthesized ones via crowdsourcing. It has 40k questions against about 2k theories. The statements are expressed in a more natural way, posing a discrepancy between DU0-DU5. It has 28k / 4k / 8k samples in the train / validation / test split respectively. In \secref{sec:generalization}, we combine it with the extensive DU3 for training, resulting in a train set containing 119k samples.

\subsection{Possible Limitations of Our Model}

Since our strategy prediction module and operations corresponding to different strategies in node prediction modules are specially designed for the current datasets, we may need to redesign some specific operations to reach the best performance, if some novel proof types are included in new datasets. But we believe our architecture will still take effect without modification. Besides, the interpretability of \ourmodel is not so strong as former works like EVR that make use of intermediate texts.

\subsection{Strategy Accuracy of IBR}
\label{app:strategy_acc}

\begin{table}[h]
\centering
\begin{tabular}{llc}
\toprule
\textbf{D}   & Cnt   & Strategy Accuracy \\ \midrule
0   & 6299  & 99.9              \\
1   & 4434  & 99.1              \\
2   & 2915  & 99.3              \\
3   & 2396  & 99.0              \\
4   & 2134  & 99.2              \\
5   & 2003  & 99.7              \\ \midrule
All & 20192 & 99.4              \\ 
\bottomrule
\end{tabular}
\caption{Strategy accuracy of IBR on test split of DU5 after training on training split of DU5.}
\label{tab:strategy_acc}
\end{table}
We provide the strategy prediction accuracy on DU5 in Table~\ref{tab:strategy_acc}. It proves that IBR is also well able to make predictions on the proof strategies. This is partly due to RoBERTa's powerful representation capability. On the other hand, there is a certain connection between the answer to the question and the strategy, and there are some common elements at the semantic representation level that can be learned together.

\subsection{Performance of EVR and \ourmodel on \emph{Fail-proof} Samples}
\label{app:evr_fail_results}

As we have discussed in \secref{sec:main_results}, EVR~\cite{evr} is not applicable for samples containing \emph{Fail-proof} proofs, because it cannot obtain proper intermediate questions to proceed correct following reasoning. Here, we compare our model with EVR on these samples in DU0-DU5, as illustrated in Table~\ref{tab:evr_fail_results}. Although EVR can achieve promising performance on answer prediction (QA) for these samples, it cannot generate any correct proof path in such cases, which have already been discussed in its original paper.

\begin{table}[t!]
\setlength{\abovecaptionskip}{0.1cm}
\setlength{\belowcaptionskip}{-0.0cm}
\setlength{\tabcolsep}{4pt}
\small
    \begin{tabular}{llrrrrrrr}
    \toprule
    & \textbf{D}   & \textbf{0}   & \textbf{1}    & \textbf{2}    & \textbf{3}   & \textbf{4}   & \textbf{5}   & \textbf{all} \\
    \midrule 
    & \textbf{Cnt} & 4365         & 2500          & 981          & 462         & 200         & 69         & 8577 \\
    \midrule
    \multirow{2}{*}{\textbf{QA}}
    & EVR &           99.7          & 99.1 & \textbf{98.9} & \textbf{99.1} & \textbf{98.5} & 100 & \textbf{99.4} \\
    & \textbf{IBR} & \textbf{100} & \textbf{99.1}          & 98.3          & 97.6         & 96.5         & \textbf{100}         & 99.3 \\
    \midrule
    \multirow{2}{*}{\textbf{PA}}
    & EVR &           0.0          & 0.0          & 0.0          & 0.0         & 0.0         & 0.0         & 0.0 \\
    & \textbf{IBR} & \textbf{99.8} & \textbf{95.0} & \textbf{89.5} & \textbf{84.4}& \textbf{65.5}& \textbf{28.9}& \textbf{95.0} \\
    \midrule
    \multirow{2}{*}{\textbf{FA}}
    & EVR &           0.0          & 0.0          & 0.0          & 0.0         & 0.0         & 0.0         & 0.0 \\
    & \textbf{IBR} & \textbf{99.8} & \textbf{95.0} & \textbf{89.5} & \textbf{84.4}& \textbf{65.5}& \textbf{28.9}& \textbf{95.0} \\
    \bottomrule
    
    \end{tabular}
\caption{The performance of EVR and \ourmodel on the partial test split of DU5 that only contains samples whose proofs strategies are \emph{Fail-proof}. }
\label{tab:evr_fail_results}
\end{table}

\subsection{Proof Generation samples}

\begin{figure*}[htbp]
\setlength{\abovecaptionskip}{0.1cm}
\setlength{\belowcaptionskip}{-0.0cm}
    \centering
    \includegraphics[width=1\linewidth]{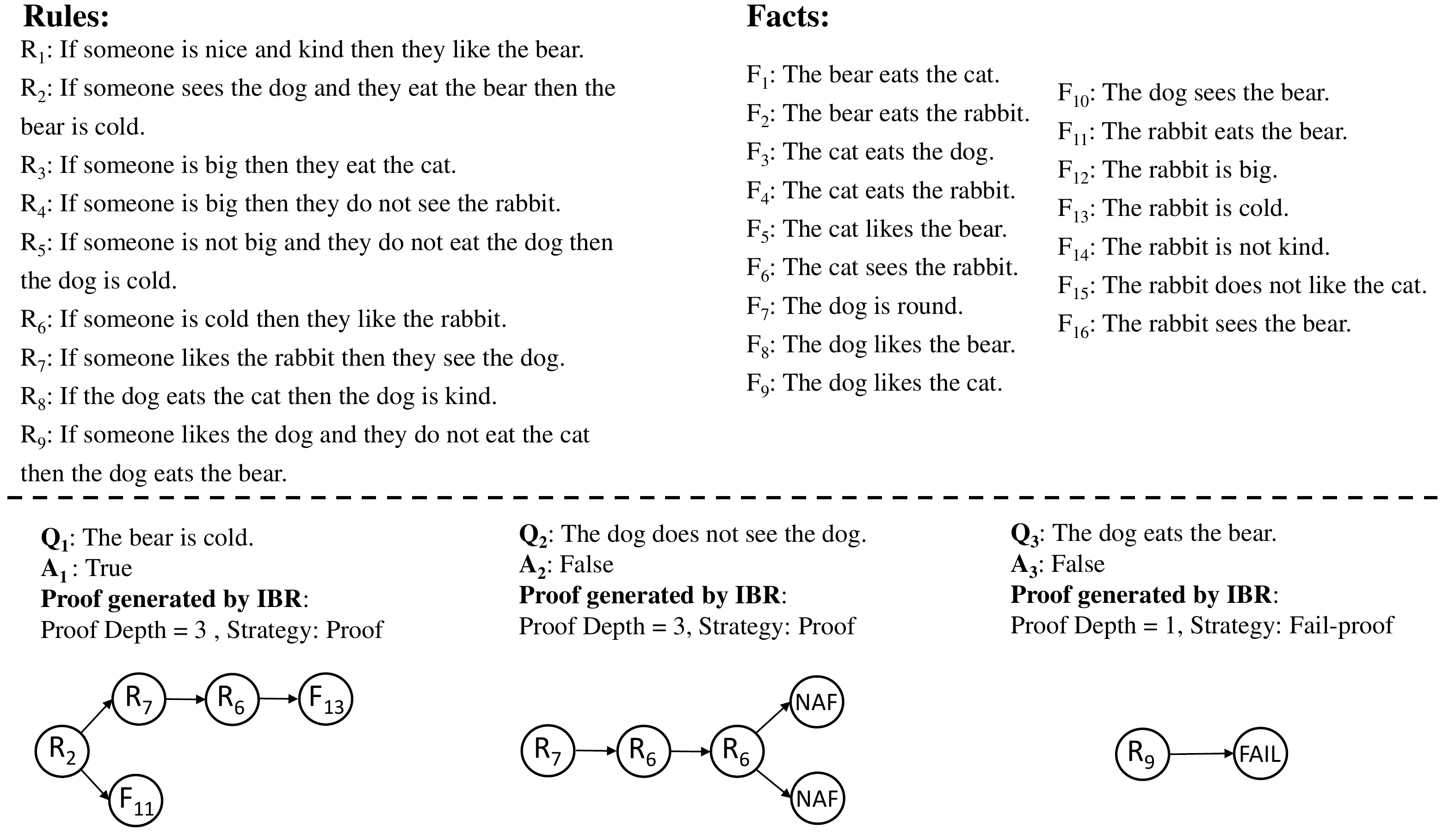}
    \caption{Some proof cases generated by \ourmodel, along with all contexts and questions, including two proof strategies, \emph{Proof} and \emph{Fail-proof}.}
    \label{fig:additional_cases}
\end{figure*}

We provide some proof generation samples in Figure~\ref{fig:additional_cases} for a better understanding of this task, where questions, all contexts, and the proof path generated by our \ourmodel are given (all consistent with the given labels).

\end{document}